\documentclass[9pt,sigconf,nonacm]{acmart}

\AtBeginDocument{%
  \providecommand\BibTeX{{%
    \normalfont B\kern-0.5em{\scshape i\kern-0.25em b}\kern-0.8em\TeX}}}

\usepackage{listings}
\usepackage{enumitem}
\usepackage{algpseudocode}
\usepackage{pgfplots}
\usepackage{changepage}
\usepackage{caption}
\usepackage{subcaption}
\usepackage{natbib}
\usepackage{balance}
\usepackage{algorithm}

\begin{document}

\newcommand{\AMcomment}[1]{\textcolor{blue}{[\textbf{AM:} #1]}}
\newcommand{\AMtext}[1]{{\textcolor{blue}{#1}}}

\title{Secure Aggregation for Federated Learning in Flower}

 \author{Kwing Hei Li}
  \affiliation{
\institution{University of Cambridge}
\country{UK}
}
 \author{Pedro Porto Buarque de Gusm\~ao}
  \affiliation{
\institution{University of Cambridge}
\country{UK}
}
 \author{Daniel J. Beutel}
  \affiliation{
\institution{University of Cambridge}
\country{UK}
}
 \author{Nicholas D. Lane}

 \affiliation{
\institution{University of Cambridge}
\country{UK}
}

 
\authorsaddresses{}
\renewcommand{\shortauthors}{K. H. Li et al.}

\ccsdesc[500]{Security and privacy~Distributed systems security}
\ccsdesc[500]{Computing methodologies~Machine learning approaches}

 \keywords{Federated Learning, Secure Aggregation,  Secure Multi-party Computation}

 \begin{abstract}
Federated Learning (FL) allows parties to learn a shared prediction model by delegating the training computation to clients and aggregating all the separately trained models on the server. To prevent private information being inferred from local models, Secure Aggregation (SA) protocols are used to ensure that the server is unable to inspect individual trained models as it aggregates them. However, current implementations of SA in FL frameworks have  limitations, including vulnerability to client dropouts or configuration difficulties.

In this paper, we present Salvia, an implementation of SA for Python users in the Flower FL framework. Based on the SecAgg(+) protocols for a semi-honest threat model, Salvia is robust against client dropouts and exposes a flexible and easy-to-use API that is compatible with various machine learning frameworks. We  show that Salvia's experimental performance is consistent with  SecAgg(+)'s theoretical computation and communication complexities.
 \end{abstract}
\maketitle

 
\section{Introduction}
Federated Learning (FL)~\cite{FL,FL2} is a recent machine learning (ML) paradigm that allows a centralized \emph{server} to compute a global model by aggregating local models trained by a set of  \emph{clients}. Though not having direct access to users' data,
a malicious server can still infer patterns of private data through inference attacks on clients' local models~\cite{ML_attack, ML_attack2, diff_privacy}.




Secure aggregation (SA)~\cite{def_sec_agg}, in general, refers to any protocol that allows a group of mutually distrustful parties, each holding a private value, to compute an aggregate value without revealing any information about their private value to each other. This is especially relevant in the context of FL as we would want the server to perform the aggregation step with SA. That way, the server cannot access clients' trained models and obtain information about their private data.  

Current implementations of SA in FL frameworks generally fall under one of two main categories:


\textbf{Trusted Execution Environment: }The data-sensitive computation for aggregating models is delegated to an isolated processing environment, which is supported by trusted hardware running parallel to the operating system, e.g.\ the Intel Software Guard Extensions~\cite{sgx}. The server is only able to inspect the final result of the computation, but not any intermediate results or inputs to the computation performed in the isolated environment. 
 
 FL frameworks such as PySyft~\cite{pysyft} and OpenFL~\cite{openfl} provide support for this kind of SA, by using a lightweight library OS, Graphene~\cite{graphene}, to integrate its code with the secure hardware. However, a ML engineer may find it difficult to configure the program files to use the hardware; Configuration steps are long and complicated, and there is a lack of documentation and examples for using these trusted hardware to perform SA in  FL. In addition, various attacks~\cite{meltdown,spectre} targeting supposedly-secure hardware have been discovered in recent years.

\textbf{Multi-party Computation: } Privacy of locally-trained models is achieved by applying techniques from cryptography, e.g. Yao's garbled circuits~\cite{yao}, homomorphic encryption~\cite{homomorphic} or secret sharing~\cite{secret_example}. Instead of relying on trusted hardware, the server  operates directly on  encrypted or masked models to calculate the aggregated result without revealing individual clients' contributions. 
 
 FL frameworks such as Crypten~\cite{Crypten} provide support for this kind of SA. Though these SA methods can be designed to expose an easy-to-use API to engineers, most implementations of multi-party computation SA protocols cannot work around dropouts, a phenomenon all too common with cross-device FL.  On top of that, these protocols often incur significant computation and communication overhead, making them infeasible in larger FL experiments.

In summary, most implementations or proposed solutions for SA in common FL frameworks have one or more of the following limitations that hinder their usability:
\begin{itemize}	[leftmargin=*]
\item Not trivial to configure and use. This is the case for most frameworks, especially those that rely on trusted execution environments for SA, like PySyft and OpenFL.
\item Dependent on certain trusted hardware and   prone to existing or future attacks on it. This is the case for all  frameworks that use trusted hardware for SA, like PySyft and OpenFL.
\item Inability to tolerate client dropouts. This is the case for most frameworks that utilize multi-party computation protocols for SA, like Crypten.
\item Computation and communication overhead  too significant to be used in larger FL settings. This is the case for many proposed multi-party computation protocols, where the protocol itself is computationally expensive~\cite{paillier}.
\end{itemize}

In this paper, we present Salvia~\cite{salvia}, an implementation of SA in the open-source FL framework Flower that aims to address each of the above limitations by:
\begin{itemize}	[leftmargin=*]
\item Exposing a flexible and easily-configurable API that works with Flower's existing Strategy abstraction.
\item Utilizing a multi-party computation protocol that does not require trusted hardware.
\item Tolerating various percentages of client dropouts while providing strong security guarantees.
\item Using a SA algorithm that has  low theoretical computation and communication complexities.
\end{itemize}

We explain Salvia's system design and implementation details in section~\ref{section:system} and section~\ref{section:overview}, respectively. We also present experiments that explore the algorithmic aspects of Salvia's computation and communication costs in section~\ref{sec:experiments}. Lastly, we discuss Salvia's current limitations  and directions for future work in section~\ref{sec:limitations}. 

\section{System Design}
\label{section:system}
We  present Salvia's design goals that address  limitations of other SA implementations, and summarize the assumptions made for our implementation. We also present Flower and SecAgg(+), which are the chosen underlying framework and SA protocol for Salvia's design, respectively.

\subsection{Design Goals}
Many frameworks provide support for SA, though there are  weaknesses in their designs that hinder engineers from using them. Based on these observations, we present five main design goals for implementing Salvia:

 \textbf{1) Usability}: Given the difficulty in using SA  for some FL frameworks, Salvia should be intuitive and easy to use. 

\textbf{2) Flexibility}: Given the complexity of FL systems, Salvia should provide a flexible API for users to configure parameters of the protocol to fit their experiment/deployment goals.

 \textbf{3) Compatibility}:  Given the robust and diverse range of existing ML frameworks, Salvia should be  compatible  with the most commonly-used ones.

 \textbf{4) Reliability}: Given that dropouts of mobile devices in FL is common, Salvia should be robust against such behaviour.

 \textbf{5) Efficiency}: Given that real-world FL is often used on large systems,  Salvia should not incur significant communication and computation overhead to the FL training.


\subsection{Assumptions}
We present the assumptions made for our implementation:

\textbf{Semi-honest threat model:} Parties cannot deviate from the protocol specification. However, corrupted parties may cooperate outside the protocol to exchange information, e.g. secret shares.

 \textbf{Liveness properties on the response time of clients:} After sending a request to a client, the server eventually receives a response or detects a disconnection from the client.

 \textbf{Secure links between clients and server:} Links are encrypted and authenticated in advance.

\subsection{Framework Selection -- Flower}
\begin{figure}[h]
  \centering
  \includegraphics[width=\linewidth]{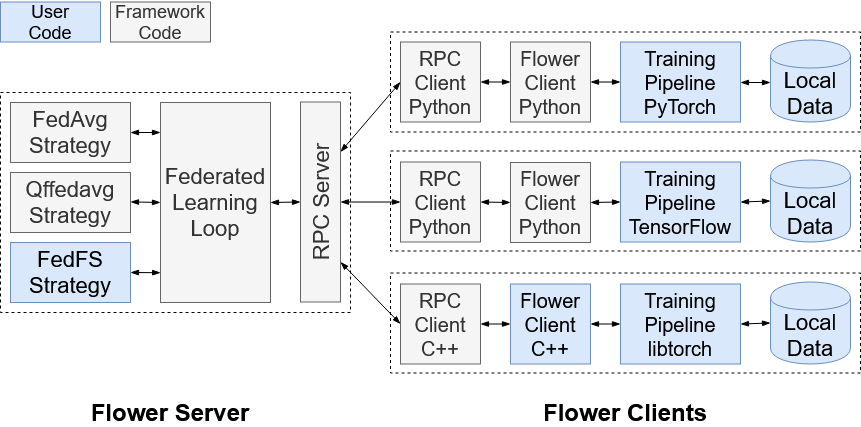}
  \caption{Flower's Basic Architecture (without Salvia)}
  
  \label{fig:flower}
\end{figure}

Flower (Figure~\ref{fig:flower})~\cite{flower} is a recent FL framework that provides higher-level abstractions enabling researchers to extend and implement FL ideas on a reliable stack. It is one of the very few frameworks that can support heterogeneous clients running on different ML frameworks (including TensorFlow~\cite{tensorflow} and PyTorch~\cite{pytorch}) and using different programming languages. Choosing Flower as Salvia's underlying FL framework allows Salvia to be compatible with many existing ML frameworks; Each client can freely choose which ML framework to use for their local training pipelines independently. Flower also has a large suite of built-in Strategies representing state-of-the-art FL algorithms for users to freely extend, modify, and use for their experiments. We also note that Flower did not have any support for SA in the past, which is a significant limitation in terms of privacy promises associated with FL.

\subsection{Protocol Selection -- SecAgg(+)}
We chose the SecAgg~\cite{SecAgg} and SecAgg+~\cite{SecAgg+} protocols for semi-honest scenarios (parties cannot deviate from the protocol)  as the base algorithm of Salvia. These multi-party computation protocols rely on cryptographic primitives to generate private masks for encrypting locally-trained models seen as a single vector of integer weights. These masks cancel  each other out when the encrypted vectors are aggregated. 

We use $n$ and $l$ to denote the number of clients participating in the protocol and the model size, respectively. We use $k$ to denote the number of other clients each client communicates with  (including itself) in the protocol, using the server as a relay. 

Like SecAgg and SecAgg+, there are five stages in  Salvia:
\vspace{1mm}

\textbf{Stage 0 -- Setup Parameters:} The server sends values of the protocol's parameters to each client.

\textbf{Stage 1 -- Ask Keys:} Each client generates private-public keys and shares the public keys via the server.

\textbf{Stage 2 -- Share Keys:} Each client generates secret shares of its private key and a randomly generated seed, and shares it with its $k$ neighbors via the server.

\textbf{Stage 3 -- Ask Vectors:} Each client creates masks for its model vector, generated with its private key and randomly generated seed. It then sends the masked vector to the server.

\textbf{Stage 4 -- Unmask Vectors:} The server asks clients to contribute secret shares they have received in the Share Keys Stage to remove the masks of the aggregated masked vector.
\vspace{1mm}

The  main difference between SecAgg  and SecAgg+ lies in the value of $k$. For the former, the value of $k$ is the same as $n$. As a result, all clients consider each other to be close neighbors, and secret shares are generated for all other clients.  For the latter, $k$ is any value smaller than $n$ (usually $O(\log n)$). Each client thus  produces shares for its closest $k$ neighbors (including itself), thus producing a $(k-1)$-connected communication graph. As both protocols differ mainly by their $k$ value \footnote{We acknowledge the fact that there exists an optimization that only works for SecAgg, but not SecAgg+ in the Unmask Vectors Stage. However, since both protocols have many features in common, we  consider SecAgg+ as a generalized version of SecAgg.} and that their underlying algorithms work similarly, we use SecAgg(+) to represent both protocols.

The computation and communication overhead complexities of Salvia are summarized in Table~\ref{table:1}.

\begin{table}[h!]
    \centering \caption{SecAgg(+)'s Overhead}
    \begin{tabular}{c|c|c}
    &  Computation   & Communication  \\
     \hline
     Server & $O(nk(k+l))$ & $O(n(k+l))$\\
     \hline
     Client & $O(k(k+l))$ & $O(k+l)$\\
    \end{tabular}
   
    \label{table:1}
\end{table}

We chose these two protocols as the base algorithm of Salvia for the following reasons:

 \textbf{Flexible API: } SecAgg(+) lends itself easily to provide a variety of parameters to be configured, allowing us to design a set of APIs that is flexible to the user's needs. Depending on the parameters used, the protocols can tolerate various percentages of corrupt users and dropouts, and is suitable for a wide range of FL scenarios of various complexities. This enables us to specifically address both our \textbf{Flexibility} and \textbf{Reliability} design goals.

\textbf{Simple Configuration: }  The execution of these protocols for both the server and clients do not rely on any special hardware support. This allows much of the logic to be directly implemented into a FL framework, decreasing the amount of configuration work a user has to set up to use Salvia. This enables us to specifically address our \textbf{Usability} design goal.

\textbf{Low Overhead: }  SecAgg's communication and computation overhead is lower than other traditional multi-party computation SA protocols. This overhead is even more significantly lowered when SecAgg+ is used, allowing a smaller value of $k$ to be used without sacrificing significant security guarantees. For example, the server's communication and computation complexities are linear to $n$. Furthermore, the client's communication and computation complexities do not increase with $n$. This enables us to specifically address our \textbf{Efficiency} design goal.

 \textbf{Clear Specifications: } SecAgg(+) served as the inspiration for many other SA multi-party computation algorithms, e.g.\ CCESA~\cite{ccesa}, TurboAgg~\cite{turbo_agg} and FastSecAgg~\cite{fast_sec_agg}. We are also not aware of other common FL frameworks that provide SA via the SecAgg(+) protocols. Implementing SecAgg(+) in a FL framework could provide valuable insights on the implementations of other similar protocols.

\section{Implementation}
\label{section:overview}
We now describe  details of implementing Salvia within Flower.

\subsection{Salvia's Architecture}
\label{subsection:arch}

\begin{figure}[h]
  \centering
  \includegraphics[width=\linewidth]{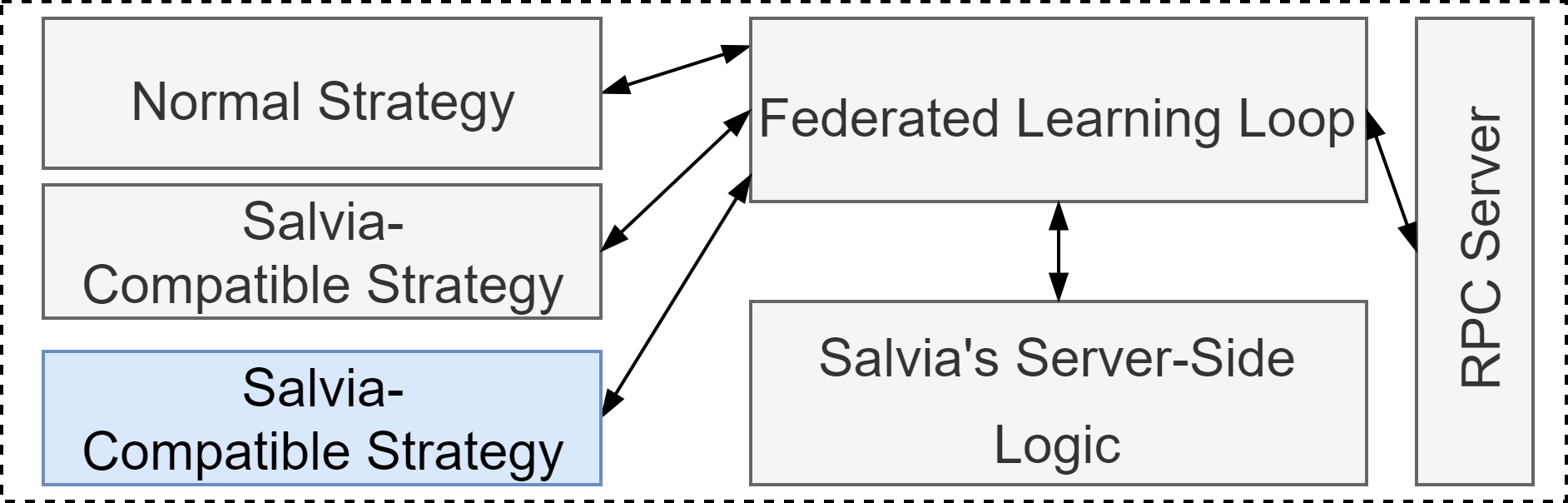}
  \caption{Server-side Architecture with Salvia\\The Salvia-compatible Strategy can be one of the standard Strategies provided by Flower (grey), or one that is implemented by the user themselves (blue).}
  \label{fig:server}
\end{figure}
\begin{figure}[h]
  \centering
  \includegraphics[width=\linewidth]{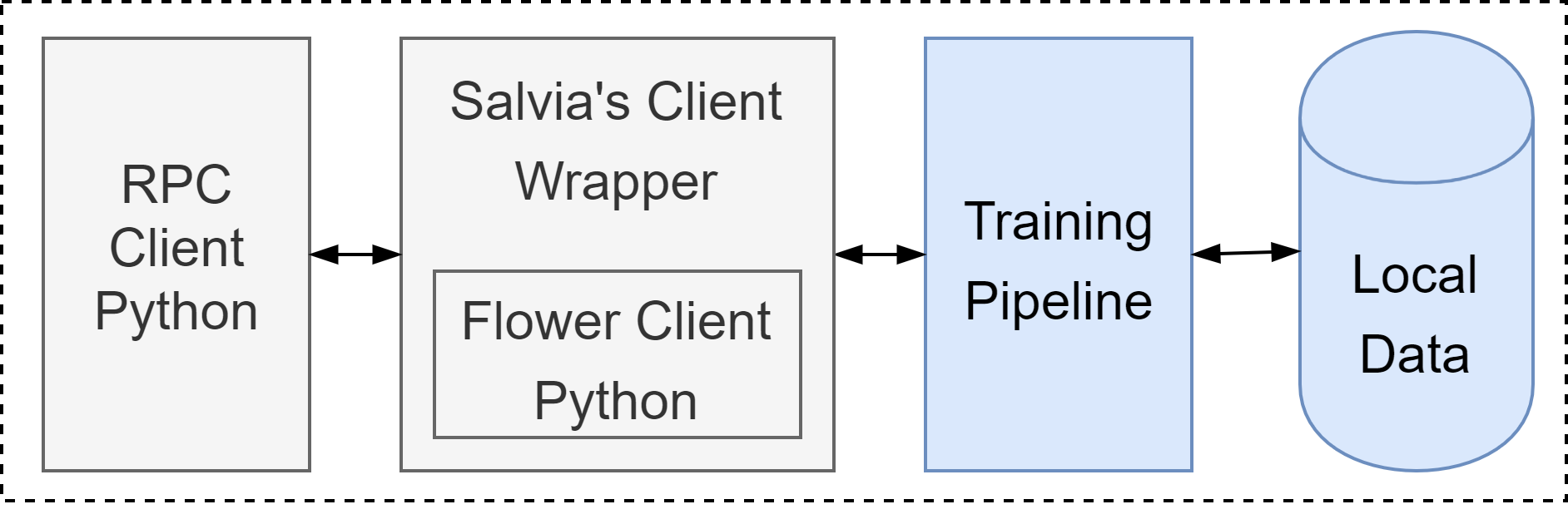}
  \caption{Client-side Architecture with Salvia}
  \label{fig:client}
\end{figure}

Salvia's architecture consists of three major components: Salvia-compatible Strategies, the server-side logic (Figure~\ref{fig:server}), and the client-side logic (Figure~\ref{fig:client}).

To use Salvia, the user must provide a Salvia-compatible Strategy that provides the configuration parameters of the protocol when starting the server, see subsection~\ref{subsection:strategy} and subsection~\ref{subsection:parameters} for details.

Salvia's server-side  logic  provides functions that are called by the FL Loop,  the heart of Flower's core architecture (Figure~\ref{fig:flower}). In a normal FL round, the loop asks the Strategy to produce configuration parameters, sends those parameters to the clients via the RPC server and client, receives the trained clients' model vectors and delegates the result aggregation to the Strategy. If the user chooses to use SA in their FL training stage, a special \textit{sec\_agg\_fit} function is executed in the  Loop, which first asks the Strategy to provide parameters that customize core aspects of the SA algorithm. 
Unlike a normal FL, the aggregation computation is performed within the Loop instead of being delegated to the Strategy. Though this limits the flexibility of the aggregation step (subsection~\ref{subsection:strategy}), this is necessary as the aggregation computation involves complicated logic to remove the  masks of the aggregated model vector, and should not be exposed to the user directly.

Salvia's  client-side logic is implemented in a wrapper class of the Flower client. Depending on the header of the messages received, the client  inspects its contents, executes the corresponding SecAgg(+) function, and responds to the server's request for each stage of the SecAgg(+) protocol.

\subsection{Cryptographic Primitives}
A suite of functions providing cryptographic primitives are used in both the server and client-side logic of Salvia. We opted not to directly provide our own implementations of these cryptographic primitives at the current stage. Instead, we make use of cryptography modules that are widely used in the Python community to limit the risk of incorrect implementation of critical cryptography functions. We discuss the cryptographic primitives used in Salvia.

\textbf{Private-Public Key Generation: }This is used to create a common seed between two clients for generating pairwise masks, and  a mutual key shared between two clients for authenticated encryption of secret shares. We used functions provided by the  Elliptic curve cryptography module in Cryptography~\cite{cryptography} for this primitive.
 
  \textbf{Authenticated Encryption: }To distribute secret shares during the Share Keys stage, the message  is encrypted by an authenticated encryption function  to guarantee confidentiality and integrity of the message. We used functions provided by the Fernet module in Cryptography for this primitive.
  
  \textbf{$t$-out-of-$n$ Secret Sharing Scheme: }Depending whether a client dropped out after the Ask Vectors Stage, the server asks all clients for a secret share they received, either from the original client's first private key, or its private seed. If at least $t$ out of $n$ shares are received by the server, then the secret is reconstructed~\cite{shamir}. Otherwise, the shares cannot give any extra information about the secret. 
    We used the Shamir's secret sharing module from the PyCryptodome~\cite{pycryptodome} library for this primitive.

   \textbf{Random Number Generator: }Each client produces a random number that acts as the seed for generating its own private mask of its model vector. 
    We used the random function provided by the Python~\cite{python} \textit{os} library   for this primitive.
    
     \textbf{Pseudo Random Number Generator: }To ensure pairs of clients generate the same pairwise masks, a pseudo-random number generator is required so that the same mask is produced among clients when given identical seeds.
   We used the standard random function in Python's standard library for this primitive.

\subsection{Federation Strategies}
\label{subsection:strategy}

In Flower, users can experiment  with state-of-the-art algorithms and modify the behavior of their FL workload through the Strategy abstraction. A Strategy customizes core aspects of the FL process, e.g.\ client sampling and  training parameters. 
 Currently, Flower provides a comprehensive suite of FL Strategies, representing the wide range of FL algorithms used by the FL community, e.g.\ FedAvg~\cite{FL} and FedProx~\cite{fed_prox}. In addition, users can extend or modify these built-in Strategies, or even implement their own. 
 We address our \textbf{Usability} and \textbf{Compatibility} design goals by allowing Salvia to be  compatible with this Strategy abstraction, i.e. users can choose Strategies easily and flexibly to be used in conjunction with Salvia.
 
 For a Strategy to be  Salvia-compatible, the Strategy must also be a subclass of the \textit{SecAggStrategy} abstract base class. This means that the user needs to provide a definition of  \textit{get\_sec\_agg\_param()}, a function called by the FL Loop at the beginning of the round to obtain a dictionary of SecAgg(+)-related parameters, see subsection~\ref{subsection:parameters}. After assigning default values for undefined parameters, the FL Loop   verifies that all parameters are valid for the SecAgg(+) protocol and passes the dictionary to clients in the Setup Parameters Stage. Since the FL Loop  automatically fills in default values for undefined parameters, the implementation of the \textit{get\_sec\_agg\_param()} could be as simple as returning an empty dictionary.
  
  Like a normal Strategy, a Salvia-compatible one provides arbitrary logic and parameters to customize the client sampling and client's model fitting process. However, as mentioned in subsection~\ref{subsection:arch}, it cannot tune the weighted  aggregation process which occurs together with the Unmask Vectors Stage in the FL Loop, see subsection~\ref{subsection:weighted_agg}.


        
 

\subsection{API Parameters}
\label{subsection:parameters}
To achieve our \textbf{Flexibility} goal, users can specify  values of Salvia-related parameters  for configuring the SA protocol via the dictionary returned by the \textit{get\_sec\_agg\_param()} function of their Strategy. These parameters are categorized into these classes:

 \textbf{Minimum Number of Clients:} To prevent the server from computing an aggregated vector of too few people (an aggregated vector from a single client in the extreme case), the user can specify the minimum number of clients required to be available by the end of the protocol. If the number of available clients drops below this limit, the server and clients  refuse to continue the protocol for security reasons.
    
    The user can control this limit through one of two ways, by providing an exact value of this limit with the parameter \textit{min\_num}, or by specifying a fraction with respect to the number of clients sampled in \textit{min\_frac}. If both are provided,  the least restrictive of the two is used.

 \textbf{Secret Sharing:} The generation and reconstruction functions of the secret sharing cryptographic primitive is controlled by two parameters provided by the user. \textit{share\_num} specifies the number of shares generated for each secret, and \textit{threshold} specifies the minimum number of shares required to reconstruct the secret. Incidentally, \textit{share\_num} also represents the value of $k$, i.e. the number of neighbors each client communicates with (including itself). Therefore, if the value of \textit{share\_num} is identical to the number of clients sampled, the protocol executed is SecAgg.
    

 \textbf{Quantization:} For masks generated by cryptographic primitives to work properly, client's locally-trained model vector must be quantized beforehand. This quantization step can be configured through the \textit{clipping\_range} and \textit{target\_range} parameters. For each real value of the trained model vector, clients first clip the value  within the range $[-$\textit{clipping\_range}, \textit{clipping\_range}$]$ and then map it to an integer between 0 and (\textit{target\_range}-1) uniformly. Similarly, an inverse translation is performed on the server's side  by converting each integer element of the quantized aggregated vector to the float type, and  mapping it to a real value in the range $[-$\textit{clipping\_range}, \textit{clipping\_range}$]$ uniformly to produce the real-valued aggregated model vector.
 
 Though information is lost after quantization which might affect the accuracy of the FL, this procedure is compulsory for the cryptographic primitives to work properly. A user can decrease the amount of information lost from quantization by providing a sufficiently large value of \textit{target\_range}.


   \textbf{Weighted Aggregation: }  \textit{max\_weights\_factor}  denotes the maximum weight that could be applied to a client's model vector for weighted aggregation. The \textit{mod\_range} parameter specifies the modulus of the aggregation computation, and also the range of values elements in client's vector masks can take, see subsection~\ref{subsection:weighted_agg}.

\subsection{Weighted Aggregation}
\label{subsection:weighted_agg}
Fundamentally, the SecAgg and SecAgg+ protocols only allow the server to calculate an unweighted average of all received model vectors. However, we often want to calculate a weighted average of all vectors, e.g.\ a client who has trained on more data should have a larger influence on the final global model than one who has trained with fewer. This is not trivial because the server cannot perform arbitrary operations on clients' masked vectors without modifying the masks themselves.

We address this issue by applying the technique described in the appendix of~\cite{SecAgg}: Salvia's client-side logic automatically multiplies the client's model vector by the amount of training data used, and appends that factor to the front of the vector, producing a modified vector to be masked. On the server's side, at the end of the Unmask Vectors Stage, the server  pops the first element of the vector, representing the sum of all  factors, and divides the rest of the vector by that value to obtain a weighted average of all input vectors.

The \textit{max\_weights\_factor} prevents clients from multiplying their vector by too large a factor that risks `overflowing' the field and affecting the aggregation result. If a client trains its model by an amount of data that exceeds  \textit{max\_weights\_factor}, the vector is only multiplied by  \textit{max\_weights\_factor} instead of the actual amount of data trained. In addition, after all Salvia-related parameters are defined, the FL Loop checks that \textit{mod\_range} is at least as large as $($\textit{max\_weights\_factor} $\times$ \textit{target\_range} $\times$ number of sampled clients $)$ so that it is impossible for the  aggregated vector to exceed the chosen modulus field.

An interesting side-effect of this design choice is that we can execute unweighted aggregation  by setting \textit{max\_weights\_factor} to $1$, as each client  automatically multiplies their vector by the same factor of value $1$.

\section{Evaluation}
\label{sec:experiments}
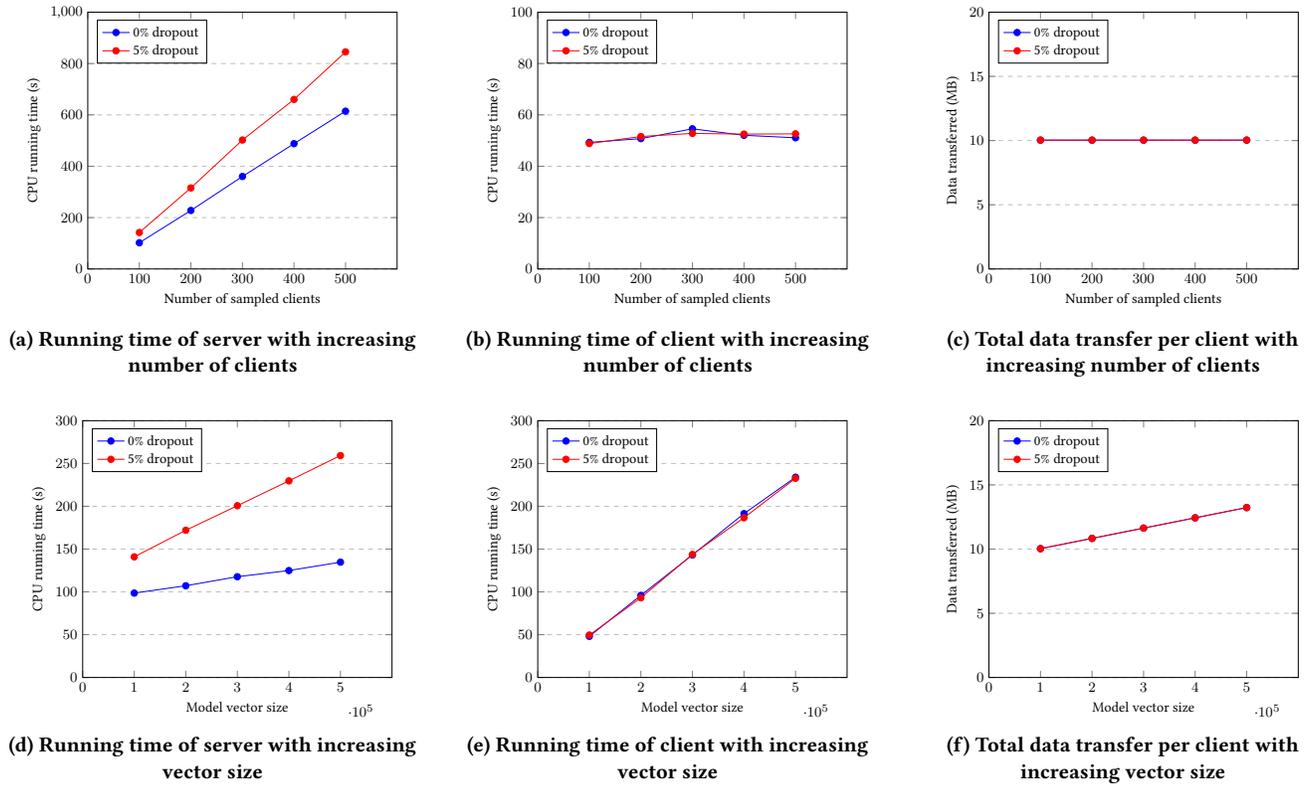
\begin{figure*}

        \centering
        \begin{subfigure}[b]{0.32\textwidth}
            \centering
            \captionsetup{justification=centering}
            \begin{tikzpicture}[scale=0.60]
\begin{axis}[align =center,
    title={},
    xlabel={Number of sampled clients},
    ylabel={CPU running time (s)},
    xlabel near ticks,
    ylabel near ticks,
    xmin=0, xmax=600,
    ymin=0, ymax=1000,
    xtick={0,100,200,300,400,500},
    ytick={0,200,400,600,800,1000},
    legend pos=north west,
    ymajorgrids=true,
    grid style=dashed,
]

\addplot[
    color=blue,
    mark=*
    ]
    coordinates {
    (100,102.1089189)(200,227.7907158)(300, 360.1610669)(400,487.8874253)(500,614.4436418)
    };

\addplot[
    color=red,
    mark=*
    ]
    coordinates {
    (100, 141.9952521)(200,315.5770481)(300,502.1805134)(400,660.1611714)(500,845.3339915)
    };
    \legend{0\% dropout,5\% dropout}
    
\end{axis}
\end{tikzpicture}
            \caption{{\small Running time of server with increasing number of clients}}    
            \label{fig:server_number}
        \end{subfigure}
        \hfill
        \begin{subfigure}[b]{0.32\textwidth}  
            \centering 
            \captionsetup{justification=centering}
            \begin{tikzpicture}[scale=0.60]
\begin{axis}[align=center,
    title={},
    xlabel={Number of sampled clients},
    ylabel={CPU running time (s)},
    xlabel near ticks,
    ylabel near ticks,
   xmin=0, xmax=600,
    ymin=0, ymax=100,
    xtick={0,100,200,300,400,500},
    ytick={0,20,40,60,80,100},
    legend pos=north west,
    ymajorgrids=true,
    grid style=dashed,
]

\addplot[
    color=blue,
    mark=*
    ]
    coordinates {
    (100,49.28518166)(200,50.76211393)(300, 54.57449959)(400,52.05518935)(500,51.09066227)
    };

\addplot[
    color=red,
    mark=*
    ]
    coordinates {
    (100, 48.84764945)(200,51.57511236)(300,52.79719244)(400,52.51929505)(500,52.61413222)};\legend{0\% dropout,5\% dropout}
    
\end{axis}
\end{tikzpicture}
            \caption{{\small  Running time of client with increasing number of clients}}    
            \label{fig:client_number}
        \end{subfigure}
        \hfill
        \begin{subfigure}[b]{0.32\textwidth}   
            \centering 
            \captionsetup{justification=centering}
            \begin{tikzpicture}[scale=0.60]
\begin{axis}[align=center,
    title={},
    xlabel={Number of sampled clients},
    ylabel={Data transferred (MB)},
    xlabel near ticks,
    ylabel near ticks,
   xmin=0, xmax=600,
    ymin=0, ymax=20,
    xtick={0,100,200,300,400,500},
    ytick={0,5,10,15,20},
    legend pos=north west,
    ymajorgrids=true,
    grid style=dashed,
]

\addplot[
    color=blue,
    mark=*
    ]
    coordinates {
    (100,10.03374636)(200,10.0338654)(300, 10.03401227)(400,10.0342582)(500,10.03440576)
    };

\addplot[
    color=red,
    mark=*
    ]
    coordinates {
    (100,10.03492441)(200,10.03504254)(300,10.03518886)(400,10.03543452)(500,10.03558192)};\legend{0\% dropout,5\% dropout}
    
\end{axis}
\end{tikzpicture}
            \caption{{\small Total data transfer per client with increasing number of clients}}    
            \label{fig:client_number_communication}
        \end{subfigure}
         \vskip\baselineskip
         \begin{subfigure}[b]{0.32\textwidth}   
            \centering 
            \captionsetup{justification=centering}
            \begin{tikzpicture}[scale=0.60]
\begin{axis}[align =center,
    title={},
    xlabel={Model vector size},
    ylabel={CPU running time (s)},
    xlabel near ticks,
    ylabel near ticks,
    xmin=0, xmax=600000,
    ymin=0, ymax=300,
    xtick={0,100000,200000,300000,400000,500000},
    ytick={0,50,100, 150,200,250,300},
    legend pos=north west,
    ymajorgrids=true,
    grid style=dashed,
]

\addplot[
    color=blue,
    mark=*
    ]
    coordinates {
    (100000,98.58514578)(200000,107.1396354)(300000, 117.7148892)(400000, 124.9209483)(500000, 134.721084)
    };

\addplot[
    color=red,
    mark=*
    ]
    coordinates {
    (100000, 140.9449628)(200000,172.0023658)(300000,200.6615239)(400000, 229.6977153)(500000, 259.2806949)
    };
    \legend{0\% dropout,5\% dropout}
    
\end{axis}
\end{tikzpicture}
            \caption{{\small Running time of server with increasing vector size}}    
            \label{fig:server_vector}
        \end{subfigure}
         \hfill
         \begin{subfigure}[b]{0.32\textwidth}   
            \centering 
            \captionsetup{justification=centering}
            \begin{tikzpicture}[scale=0.60]
\begin{axis}[align =center,
    title={},
    xlabel={Model vector size},
    ylabel={CPU running time (s)},
    xmin=0, xmax=600000,
    ymin=0, ymax=300,
    xlabel near ticks,
    ylabel near ticks,
    xtick={0,100000,200000,300000,400000,500000},
    ytick={0,50,100, 150,200,250,300},
    legend pos=north west,
    ymajorgrids=true,
    grid style=dashed,
]

\addplot[
    color=blue,
    mark=*
    ]
    coordinates {
    (100000,48.02484683)(200000,95.90823009)(300000, 143.1134221)(400000, 191.4132502)(500000, 233.9891518)
    };

\addplot[
    color=red,
    mark=*
    ]
    coordinates {
    (100000, 49.58145018)(200000,93.06931294)(300000,143.7644481)(400000, 186.59506)(500000, 232.7145917)
    };
    \legend{0\% dropout,5\% dropout}
    
\end{axis}
\end{tikzpicture}
            \caption{{\small Running time of client with increasing vector size}}    
            \label{fig:client_vector}
        \end{subfigure}
        \hfill
        \begin{subfigure}[b]{0.32\textwidth}   
            \centering 
            \captionsetup{justification=centering}
            \begin{tikzpicture}[scale=0.60]
\begin{axis}[align =center,
    title={},
    xlabel={Model vector size},
    ylabel={Data transferred (MB)},
    xlabel near ticks,
    ylabel near ticks,
    xmin=0, xmax=600000,
    ymin=0, ymax=20,
    xtick={0,100000,200000,300000,400000,500000},
    ytick={0,5,10, 15,20},
    legend pos=north west,
    ymajorgrids=true,
    grid style=dashed,
]

\addplot[
    color=blue,
    mark=*
    ]
    coordinates {
    (100000,10.03374636)(200000,10.83374636)(300000, 11.63375036)(400000,12.43375036)(500000,13.23375036)
    };

\addplot[
    color=red,
    mark=*
    ]
    coordinates {
    (100000,10.03492441)(200000,10.83492441)(300000,11.63492841)(400000, 12.43492841)(500000, 13.23492841)
    };
    \legend{0\% dropout,5\% dropout}
    
\end{axis}
\end{tikzpicture}
            \caption{{\small Total data transfer per client  with increasing vector size}}    
            \label{fig:client_vector_communication}
        \end{subfigure}
        \caption{Experiment Results}
        \label{fig:plot}
    \end{figure*}

To verify that Salvia's behavior matches our  expectations in theoretical complexity, we evaluate the changes of Salvia's computation and communication overhead with the number of clients or the model vector size. We ran our FL simulations on a Linux workstation with an Intel Xeon E-2136 CPU (3.30GHz), with 256 GB of RAM. In our simulations, all entries of our local vectors were of size 24 bits. We ignore communication latency in our simulations. Moreover, all dropouts simulated happened after stage 2, i.e. Share Keys Stage. This is because this would impose the most significant overhead as the server not only needs to regenerate their secret, but also compute their pairwise masks generated between their neighbors. 

For  our simulations, \textit{share\_num} and \textit{threshold} were set to 51 and 26, respectively. These parameters were  chosen to reference SecAgg+'s proven correctness and  security guarantees, where we can tolerate up to $5\%$  dropouts and $5\%$ corrupted clients with correctness holding with probability $1-2^{-20}$ and security holding with probability $1-2^{-40}$. Though not shown in our following simulations, users can choose larger values for \textit{share\_num} and \textit{threshold} to tolerate higher percentages of dropouts and corrupted clients. 

We observe that all our experiment results are consistent with the theoretical computation  and communication overhead complexities for SecAgg(+) from Table~\ref{table:1}, thus achieving our \textbf{Efficiency}  design goal.

\textbf{Varying the number of clients: }
We measured the CPU running times of the server and a client, and the total data transfer per client, as the number of sampled clients increases. Fixing the model vector size to 100k entries, we plotted the results measured through sampling 100 clients to sampling up to 500 clients in Figure~\ref{fig:server_number}, \ref{fig:client_number}, and \ref{fig:client_number_communication}. We also measured how the performance would change after client dropouts by repeating the  experiments with a $5\%$  dropouts. 

We observe that the server's running time increases linearly with the number of sampled clients, which matches the expected computation cost's complexity as the server repeats the same operations for each available client, e.g.\ reconstructing secrets and generating masks. The server's running time increases whenever there are $5\%$ clients dropping out, as the server has to perform extra computations to calculate all $k$ pairwise masks for each client dropped out in the Unmask Vectors Stage. On the other hand, the client's running time remains constant with the number of sampled clients regardless whether there are  dropouts. This is because each client only communicates with $k$ other neighbors and the actual number of sampled clients does not affect the client-side logic. 
Lastly, we note that the total data transferred per client remains unchanged as each client is only communicating with exactly $k$ neighbors regardless of the total number of clients and dropouts.

\textbf{Varying vector size: }
We  measured the CPU running times of the server and a client, and the total data transfer per client, as the model vector size increases. Fixing the number of sampled clients to 100, we plotted the results measured through aggregating a vector of size 100k entries to aggregating one of size 500k entries in Figure~\ref{fig:server_vector}, \ref{fig:client_vector}, and \ref{fig:client_vector_communication}. Like before, we repeated our experiments with a $5\%$  dropouts. 

We observe that both the server's and clients' running times increase linearly with the model vector size. This matches both expected computation cost's complexities because all computation costs involving model vectors are linear to the vectors' sizes, e.g.\ generating masks and unmasking vectors.  In addition, like before, the server's running time increases and the client's running time remains unchanged when there are $5\%$ clients dropping out. We also observe that the total data transferred per client increases linearly with the model vector size as expected, because each client sends the vector to the server for aggregation.


\section{Limitations and Future Work}
\label{sec:limitations}
We recognize a few limitations of our current work and we  point out  future work to address these issues.

At this moment, the secret sharing scheme function is one of the most significant bottlenecks on Salvia's computation cost. In the future, we wish to provide our own implementation of the secret sharing mechanism that is more efficient while not sacrificing security guarantees. 

In addition, our support for SA on Flower is only limited to Python users. We aim to rewrite cryptographic functions that are specific to the Python library, such that it can be used in FL scenarios where clients are running under different programming languages on their backends, e.g.\ C++. 


Lastly, we would want to allow users to customize more aspects of their FL experiments. For example, Salvia fixes clients' weights factor to be the amount of training data used. In addition, users cannot customize the quantization step. We hope to extend our implementation to allow users to provide arbitrary logic specifying how clients compute the weights factor and how quantization is performed via the Salvia-compatible Strategy.


\section{Conclusion}

We presented Salvia, an open-source implementation of SA based on the SecAgg(+) protocols in the FL framework Flower. Leveraging Flower's ML framework-agnostic property, Salvia is compatible with a diverse range of ML frameworks. We explained how Salvia's API provides a set of flexible and easily-configurable parameters and how it works together with Flower's Strategy abstraction.   We  showed that Salvia can handle client dropouts, and its performance is consistent with SecAgg(+)'s theoretical computation and communication complexities. Future work includes improving the efficiency of the secret-sharing functions, extending Salvia for clients running in other programming languages, and providing support for configuring the weights factor  via the Strategy abstraction.
\begin{acks}
This work was supported by the UK’s Engineering and Physical Sciences Research Council (EPSRC) with grant EP/S001530/1 (the MOA project) and the European Research Council (ERC) via the REDIAL project (Grant Agreement ID: 805194). We would also like to thank Prof.\ Mycroft and the anonymous reviewers for helpful discussions and suggestions. 
\end{acks}

 \bibliographystyle{samples/ACM-Reference-Format}
  \interlinepenalty=10000
 \bibliography{main}










\end{document}